\documentclass{article}

\usepackage{microtype}
\usepackage{graphicx}
\usepackage{subfigure}
\usepackage{booktabs} 
\usepackage{hyperref}
\usepackage{url}
\usepackage{multirow}   
\usepackage{array}      
\usepackage{caption}    
\usepackage{adjustbox}  
\usepackage{siunitx}    

\usepackage[accepted]{icml2025}

\makeatletter
\renewcommand{\Notice@String}{Preprint.}
\makeatother

\usepackage{amsmath}
\usepackage{amssymb}
\usepackage{mathtools}
\usepackage{amsthm}

\usepackage[capitalize,noabbrev]{cleveref}

\theoremstyle{plain}

\theoremstyle{definition}

\theoremstyle{remark}

\usepackage[textsize=tiny]{todonotes}

\icmltitlerunning{Improving Recursive Transformers with Mixture of LoRAs}

\begin{document}

\twocolumn[
\icmltitle{Improving Recursive Transformers with Mixture of LoRAs}



\icmlsetsymbol{equal}{*}

\begin{icmlauthorlist}
\icmlauthor{Mohammadmahdi Nouriborji}{nlpie}
\icmlauthor{Morteza Rohanian}{zurich}
\icmlauthor{Omid Rohanian}{nlpie,oxford}

\end{icmlauthorlist}

\icmlaffiliation{nlpie}{NLPIE Research, UK}
\icmlaffiliation{zurich}{University of Zurich, Switzerland}
\icmlaffiliation{oxford}{Department of Engineering Science, University of Oxford, UK}

\icmlcorrespondingauthor{Mohammadmahdi Nouriborji}{m.nouriborji@nlpie.com}
\icmlcorrespondingauthor{Omid Rohanian}{omid.rohanian@nlpie.com}

\icmlkeywords{Machine Learning, ICML}

\vskip 0.3in
]



\printAffiliationsAndNotice{}  

\begin{abstract}
Parameter sharing in recursive transformers reduces model size but collapses layer-wise expressivity. We propose Mixture of LoRAs (MoL), a lightweight mechanism that uses low-rank experts to simulate a Mixture-of-Experts layer. By replacing feed-forward networks (FFNs) in the shared transformer with these conditional modules, we effectively restore the expressivity lost due to parameter sharing. We pretrain a modernised recursive architecture, ModernALBERT, integrating rotary embeddings, GeGLU, FlashAttention, and a distillation-based initialisation. Across GLUE, SQuAD-v2, and BEIR, ModernALBERT (50M–120M) achieves state-of-the-art performance among compact models and surpasses larger fully parameterised baselines. We also propose an expert-merging procedure that compresses MoL into a single adapter at inference while preserving accuracy, enabling efficient deployment. Our results show that conditional computation effectively restores the expressivity lost under aggressive parameter sharing in recursive transformers. \footnote{Our models are available at \url{https://huggingface.co/collections/nlpie/modern-recursive-transformers}}
\end{abstract}

\section{Introduction}

\begin{figure}[t]
    \centering
    \includegraphics[width=0.5\textwidth]{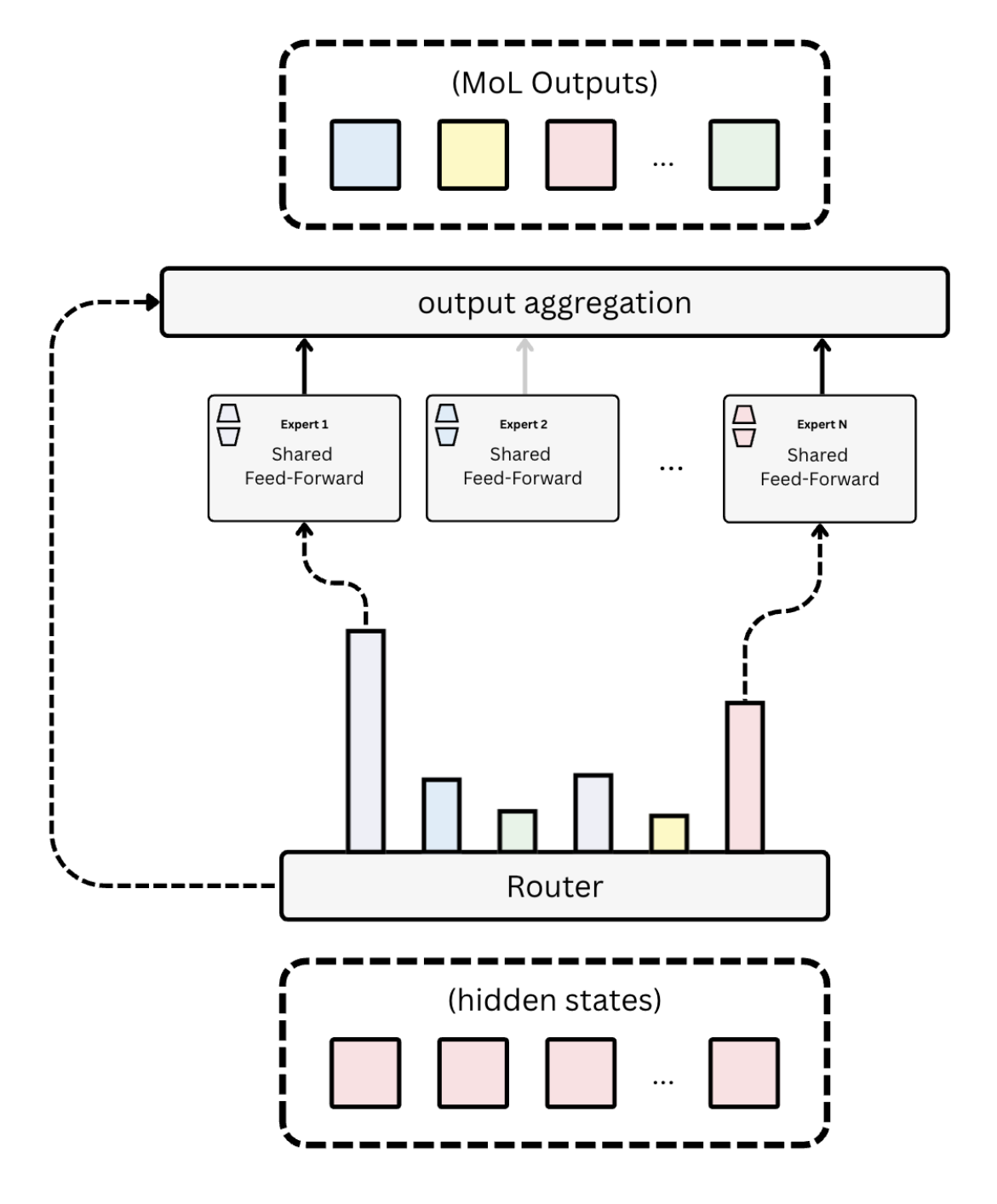}
    \caption{Structure of the Mixture of LoRAs (MoL) layer. Each MoL layer combines a shared feed-forward network (FFN) with several LoRA experts, allowing conditional computation without significantly increasing parameters.}
    \label{fig:mol}
\end{figure}

Large language models (LLMs) power modern NLP and deliver strong reasoning, understanding, and generation performance. Their large parameter counts and high memory use, however, make training, fine-tuning, and downstream adaptation computationally expensive \citep{vaswani2017attention, brown2020language, sanh2020distilbertdistilledversionbert}.

Parameter sharing reduces the memory footprint of transformer models by applying the same weights across layers instead of learning distinct parameters. ALBERT demonstrates this approach by using extensive cross-layer sharing to build compact yet strong language models \citep{lan2019albert}. Analyses show, however, that matching the performance of fully parameterised transformers can force the shared layer to grow wider, increasing its hidden dimension and computational cost \citep{lan2019albert}. This trade-off undermines efficiency and motivates approaches that preserve expressivity while minimising both parameters and computation.

Recent advances in large-scale models show the effectiveness of Mixture-of-Experts (MoE) architectures in enhancing transformer expressivity. By activating a subset of specialised expert modules for each input, MoE models significantly expand a transformer's representational capacity and deliver strong performance across diverse tasks \citep{shazeer2017mixture, fedus2021switch, lepikhin2020gshard}. Despite this success, MoE architectures introduce substantial parameter growth, making direct application to shared or lightweight architectures incompatible with the goal of parameter efficiency.

Existing approaches attempt to mitigate the expressivity loss caused by parameter sharing in three main ways. (i) Static adapters (e.g., bottleneck adapters or LoRAs) add depth-specific parameters but do not adapt at the token level and do not modify the shared FFN itself. (ii) Mixture-of-Adapters (MoA) and LoRA-based MoE variants introduce dynamic experts after the FFN, leaving its internal transformation unchanged. (iii) Relaxed Recursive Transformers (RRT) attach static LoRA modules to a shared block, partially restoring depth-specific behaviour but lacking token-dependent routing. Crucially, none of these designs introduce conditional computation inside the shared FFN during pretraining. In this work, we explore whether such internal modulation can more effectively restore the expressivity lost due to parameter sharing.

To address these limitations, we introduce ModernALBERT, a recursive transformer that restores expressivity through Mixture of LoRAs (MoL), a mechanism for conditional computation within the shared architecture. We replace select FFNs in the shared transformer with MoL modules, embedding a small pool of LoRA experts directly into the network. By routing tokens to a sparse subset of these experts, MoL allows the shared backbone to exhibit distinct behaviours for different inputs. This approach reestablishes the layer-wise functional diversity that parameter sharing typically eliminates, all while preserving strict parameter efficiency~\citep{hu2021lora}.

We integrate several modern architectural components that improve training stability and runtime efficiency: gated GELU-style feed-forward networks (e.g., GeGLU) \citep{shazeer2020glu}, rotary position embeddings \citep[RoPE;][]{su2021roformer}, and I/O-aware attention implementations such as FlashAttention \citep{dao2022flashattention}. Empirical results demonstrate that ModernALBERT achieves state-of-the-art performance among compact transformer models, outperforming prior baselines on GLUE \citep{wang2019gluemultitaskbenchmarkanalysis}, SQuAD \citep{rajpurkar2018knowdontknowunanswerable}, and BEIR \citep{thakur2021beirheterogenousbenchmarkzeroshot} benchmarks. Our key contributions include:
\begin{itemize}
    \item We introduce Mixture of LoRAs (MoL), a conditional-computation layer that integrates low-rank experts into a shared FFN, restoring expressivity in recursive transformers.
    \item We build ModernALBERT, a family of compact recursive models that integrate RoPE, GeGLU, and FlashAttention, and we release all code and training checkpoints.
    \item We provide model variants ranging from 50M to 120M parameters and evaluate them extensively across GLUE \citep{wang2019gluemultitaskbenchmarkanalysis}, SQuAD \citep{rajpurkar2018knowdontknowunanswerable}, and BEIR \citep{thakur2021beirheterogenousbenchmarkzeroshot}.
    \item We show that MoL consistently improves performance over mixture-of-adapters and relaxed recursive transformers in controlled ablations, and we introduce a merging strategy that compresses experts into a single adapter for efficient deployment.
\end{itemize}

\section{Related Works}
Parameter-efficient language models were significantly advanced by ALBERT, which introduced factorised embedding parameterisation and aggressive cross-layer parameter sharing to dramatically reduce the memory footprint compared to models like BERT \citep{lan2019albert}. However, this strict static parameter tying, while efficient, introduces a trade-off by limiting overall expressivity and capacity. Early recursive architectures, such as the Universal Transformer \citep{dehghani2018universal}, also explored layer tying, but noted that matching the performance of fully parameterised models often required wider layers, increasing computational cost. This highlights a central challenge: static parameter sharing saves memory, but can create a performance bottleneck.

\begin{figure*}[t]
    \centering
    \includegraphics[width=\textwidth]{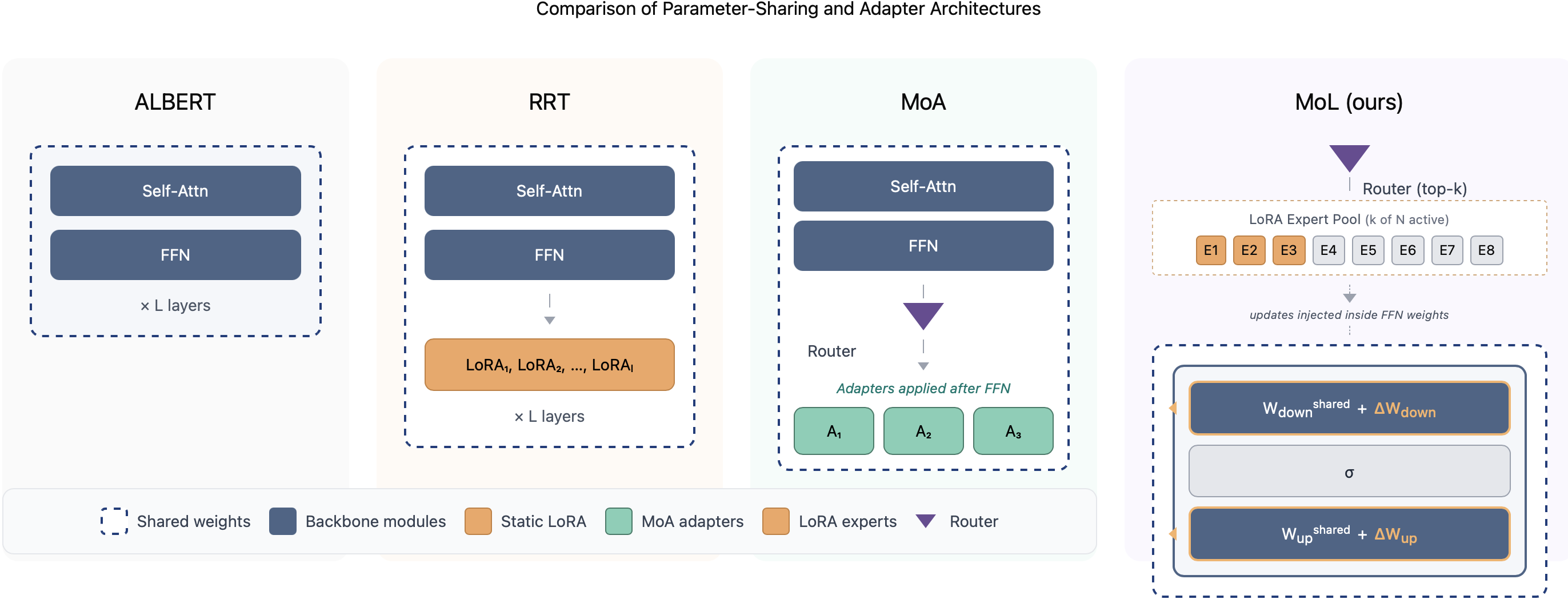}
    \caption{Comparison of parameter-sharing and adapter architectures.
    ALBERT shares a single self-attention and FFN block across all layers without adaptation.
    RRT augments the shared block with depth-specific static LoRA adapters.
    MoA keeps the shared backbone fixed and uses a router to select adapters applied after the FFN output.
    In contrast, MoL (ours) routes among a pool of LoRA experts whose low-rank updates are injected directly into the shared FFN weights, enabling token-conditional modulation while preserving parameter sharing in the backbone.}
    \label{fig:architecture-comparison}
\end{figure*}

A parallel and orthogonal approach to scaling models efficiently is MoE \citep{shazeer2017mixture, lepikhin2020gshard, fedus2021switch}. MoE models, such as the Switch Transformer, increase model capacity without a proportional rise in computation by routing each token to a sparse subset of experts. Conditional computation has also been adapted for parameter-efficient fine-tuning (PEFT), where experts are often small plug-in modules such as LoRA adapters \citep{hu2021lora, houlsby2019parameterefficienttransferlearningnlp}. Mixture-of-Adapters approaches (e.g., MoLE, MixLoRA, AdaMix, Adapter Fusion) have proven effective for multi-task fine-tuning \citep{mole2024, mixlora2024, adaptermix2023, adapterfusion2020, moa2024}. However, these frameworks are predominantly designed to be applied on top of existing models, rather than being integral to the base architecture during pretraining.

Beyond modularity, modern architectural improvements have shown measurable gains in both efficiency and performance. Enhancements such as rotary positional embeddings, gated feed-forward networks, and I/O-optimised attention implementations reduce training overhead while improving model quality. ModernBERT demonstrates that these design choices can yield substantial performance gains over standard transformer baselines without increasing computational cost \citep{warner2024modernbert}.

Recent work has explored the combination of parameter sharing with lightweight, layer-specific modules. Relaxed Recursive Transformers introduce static LoRA modules into a recursive shared block, demonstrating that depth-specific adapters can partially recover performance lost due to strict parameter sharing \citep{bae2024effective}. Similarly, MiniALBERT employs bottleneck adapters to enhance the expressivity of shared backbones, showing that compact transformers can achieve strong performance when combined with targeted, low-cost modules \citep{nouriborji2022minialbert}. While these approaches effectively augment parameter-shared models, integrating dynamic conditional computation into a compact backbone remains an open direction.

ModernALBERT addresses this limitation by introducing an architecture that combines the parameter efficiency of ALBERT-style recursive sharing with the adaptive expressivity of token-conditional expert modulation.

\section{ModernALBERT}
\textbf{ModernALBERT} is a compact transformer architecture that combines recursive parameter sharing with lightweight, parameter-efficient modules to enhance representational capacity. We outline how recursive weight sharing reduces model size while preserving effective depth, then introduce the \textbf{Mixture of LoRAs (MoL)}, which injects flexibility into shared layers through MoE-style low-rank experts. We also describe several auxiliary techniques that further stabilise training and improve performance across model sizes: knowledge distillation, improved initialisation, and architectural refinements.

\subsection{Parameter-Sharing through Recursions}
Fully-parameterised transformers, such as BERT, have demonstrated strong performance across a wide range of benchmarks and tasks. A standard transformer layer consists of a multi-head self-attention (MHA) module followed by a feed-forward network (FFN), with residual connections and layer normalisation (LN) applied around each sub-layer \citep{nguyen2019transformers}. Mathematically, the $i$-th layer can be expressed as:

\begin{align}
    h_i^{\text{att}} &= h_{i-1} + \text{MHA}(\text{LN}(h_{i-1}); \theta_i) \\
    h_i &= h_i^{\text{att}} + \text{FFN}(\text{LN}(h_i^{\text{att}}); \phi_i)
\end{align}

For brevity, let the output of the full layer be:
\begin{align}
    h_i = \Psi(h_{i-1}; \theta_i, \phi_i),
\end{align}
where $\theta_i$ and $\phi_i$ denote the parameters of the attention and feed-forward modules for the $i$-th layer, respectively. Note that in the fully-parameterised formulation, each layer has its own distinct parameters, making deep models costly and hard to deploy. Let the hidden dimension be $d$ and the FFN expansion rate be $4$; then for a transformer of depth $N$, the parameter count is approximately:

\begin{align}
    |\Theta| \approx N \Big( 4d^2 + 2\times(4d^2) \Big) = 12 N d^2    
\end{align}

To address the high parameter cost of fully-parameterised transformers, parameters can be shared across multiple layers, allowing models to maintain depth while reducing the number of unique parameters. In a recursive transformer with group size $G$ and total depth $K \cdot G$, the layers are divided into $K$ groups, where all layers within a group share parameters. Denoting the shared attention and feed-forward parameters for group $g$ as $\theta^{(g)}$ and $\phi^{(g)}$, the $i$-th layer (with $i$ and $g$ starting from 1) can be compactly written as:

\begin{align}
    h_i = \Psi(h_{i-1}; \theta^{(\lceil i/G \rceil)}, \phi^{(\lceil i/G \rceil)})   
\end{align}

In the extreme case of full parameter sharing (as in ALBERT), all layers share the same parameters, i.e., $\theta^{(g)} = \theta$ and $\phi^{(g)} = \phi$ for all layers. More generally, in a recursive transformer with total depth $K \cdot G$ and $K$ groups of size $G$, the total number of unique parameters is approximately

\begin{align}
    |\Theta| \approx K \cdot (4 d^2 + 8 d^2) = 12 K d^2
\end{align}

Compared to a fully-parameterised transformer with $N = K \cdot G$ layers, the recursive transformer reduces the number of unique parameters by a factor of $K/N$:

\begin{align}
    \frac{|\Theta|}{|\Theta_{\text{full}}|} \approx \frac{12 K d^2}{12 N d^2} = \frac{K}{N} = \frac{1}{G}
\end{align}

\begin{table*}[t!]
\centering
\small
\begin{tabular}{lccccc}
\toprule
Variant & Layers & Groups & MoL Groups & Hidden Dim & Intermediate Dim / Expert Dim \\
\midrule
Tiny & 14 & 7 & 6, 7 & 768 & 1152 / 2624 \\
Medium & 12 & 3 & 3 & 1024 & 2624 / 4096 \\
Base & 16 & 4 & 3,4 & 1024 & 2624 / 4096 \\
Large & 24 & 6 & 3,4,5,6 & 1024 & 2624 / 4096 \\
\bottomrule
\end{tabular}
\caption{Comparison of ModernALBERT model variants. MoL layers replace the shared FFNs at the indicated positions, each with 8 experts and top-2 routing.}
\label{tab:variants}
\end{table*}

\subsection{Improved Flexibility through Mixture of LoRAs}

We address the loss of representational capacity from weight sharing by integrating Mixture-of-Experts (MoE) layers, a mechanism that has proven highly effective in recent large language models. A standard MoE layer consists of a router and $E$ feed-forward modules called experts \citep{fedus2021switch}. For a given input token representation $h$, the router assigns a probability $p_i(h)$ to each expert $i$, and the output of the MoE layer is a weighted sum of the expert outputs:

\begin{align}
\text{MoE}(h) = \sum_{i=1}^{E} p_i(h) \cdot \text{FFN}_i(h)
\end{align}

where $p_i(h)$ is the routing weight for expert $i$ and $\text{FFN}_i$ is the feed-forward network corresponding to expert $i$. This allows different tokens to leverage different subsets of experts, effectively enabling conditional computation and increasing the model's flexibility.

Although MoE layers are highly effective, they come at the cost of significantly increasing the total number of parameters, which goes against our goal of building compact, efficient models. To retain the benefits of MoE while keeping the parameter count low, we adopt LoRA modules \citep{hu2021lora} to mimic the behavior of a mixture of experts. In this setup, a shared feed-forward network is augmented with multiple lightweight LoRA adapters, which act as distinct experts and enable conditional computation.

A standard LoRA module replaces a weight matrix $W \in \mathbb{R}^{d_\text{in} \times d_\text{out}}$ with a low-rank update:

\begin{align}
W' = W + \frac{\alpha}{r} AB
\end{align}

where $A \in \mathbb{R}^{d_\text{in} \times r}$ and $B \in \mathbb{R}^{r \times d_\text{out}}$  are trainable matrices of rank $r \ll d_\text{in}, d_\text{out}$, drastically reducing the number of additional parameters.

Building on this idea, we define a shared feed-forward network $\text{FFN}_\text{shared}$ with parameters $\theta = \{W_\text{down}, W_\text{up}\}$, and augment it with $E$ LoRA modules $\psi_i = \{A_{i,1}, B_{i,1}, A_{i,2}, B_{i,2}\}$ for each expert $i$. Given an input $h$, the FFN modified by the $i$-th LoRA expert is:

\begin{multline}
\text{FFN}(h;\theta,\psi_i) = \\
\big(W_{\mathrm{up}} + \tfrac{\alpha}{r} A_{i,2} B_{i,2} \big)\,
\sigma\!\Big(\big(W_{\mathrm{down}} + \tfrac{\alpha}{r} A_{i,1} B_{i,1}\big) h\Big)
\end{multline}

where $r$ is the LoRA rank, $\alpha$ is a scaling factor, and $\sigma$ is the activation function. Finally, the output of the Mixture of LoRAs (MoL) layer is computed by selecting experts via routing weights $p_i(h)$:

\begin{align}
\text{MoL}(h) = \sum_{i=1}^{E} p_i(h) \cdot \text{FFN}(h;\theta,\psi_{i})
\end{align}

We adopt a sparse top-2 routing strategy, selecting only the two experts with the highest routing weights for each token to further improve efficiency. Let $S_2(h)$ denote the indices of the top-2 experts for input $h$, then the sparse MoL output is:

\begin{align}
\text{MoL}_{\text{top-2}}(h) = \sum_{i \in S_2(h)} p_i(h) \cdot \text{FFN}(h;\theta,\psi_i)
\end{align}

where $p_i(h)$ is re-normalised over the selected top-2 experts to sum to 1.

\subsection{Effects of Knowledge Distillation and Initialisations}
We found that initialising the parameters of ModernALBERT from a fully-parameterised model, in our case ModernBERT, significantly improves performance. Following the initialisation technique proposed in Relaxed Recursive Transformers \citep{bae2024effective}, we map the parameters from the fully-parameterised model to the shared layers of ModernALBERT to provide a strong starting point for training. In addition to this, we apply knowledge distillation \citep{hinton2015distillingknowledgeneuralnetwork, sanh2020distilbertdistilledversionbert} from ModernBERT, using its predictions as soft targets to further guide the training of ModernALBERT. This combination of informed initialisation and distillation is critical for data efficiency, allowing the model to converge rapidly despite reduced training time and a pre-training budget of only 30B tokens (compared to the 1.7T tokens used for ModernBERT).

\subsection{Training Settings}
We adopt a multi-stage pre-training curriculum to maximise data efficiency within our limited token budget. The training process is divided into two distinct phases. We first warm up the model on the RedPajamas-1T dataset \citep{together2023redpajamas} for 20,000 to 30,000 steps. Following this initialisation phase, we transition to the RefinedWeb dataset \citep{penedo2023refinedwebdatasetfalconllm} for the remainder of the training, continuing for an additional 70,000 to 80,000 steps.

Throughout both stages, we maintain a fixed global batch size of 384 and a maximum sequence length of 1024 tokens. Optimisation is performed using the AdamW optimiser \citep{kingma2017adammethodstochasticoptimization}. We employ a linear warmup schedule, peaking at a learning rate of $5 \times 10^{-4}$ or $5 \times 10^{-5}$, followed by a linear decay. This regimen allows ModernALBERT to effectively adapt the distilled knowledge from the teacher model while refining the shared parameters and LoRA modules on high-quality web data.

\subsection{ModernALBERT Variants}

In addition to parameter-sharing and MoL, ModernALBERT incorporates several architectural improvements to enhance training stability, efficiency, and expressivity. We adopt pre-layer normalisation (Pre-Norm) \citep{nguyen2019transformers} and GeGLU activations \citep{shazeer2020glu} in the feed-forward networks to improve gradient flow and representation capacity. To further improve efficiency, we use rotary embeddings \citep{su2021roformer} in the attention mechanism and FlashAttention \citep{dao2022flashattention} for fast and memory-efficient attention computation. 

The MoL layers are placed at the end of each group by replacing the shared FFN with a Mixture of LoRAs layer. Each MoL consists of 8 experts with top-2 routing, except the tiny model, which has 4 experts and a top-1 routing, enabling conditional computation and enhanced flexibility without significant parameter overhead. 

We design four variants of ModernALBERT (Table \ref{tab:variants}) to target different model scales. This design balances model depth, parameter efficiency, and representational flexibility across different scales, enabling ModernALBERT to achieve high performance while maintaining compactness and efficiency.

\begin{table*}[t!]
\centering
\resizebox{\textwidth}{!}{%
\begin{tabular}{lcc
                cc
                ccc
                ccc
                c}
\toprule
\multirow{2}{*}{\textbf{Model}} & \multirow{2}{*}{\textbf{Params}} & \multirow{2}{*}{\textbf{Seq.}} 
& \multicolumn{2}{c}{\textbf{Single Sentence}} 
& \multicolumn{3}{c}{\textbf{Paraphrase and Similarity}} 
& \multicolumn{3}{c}{\textbf{Natural Language Inference}} 
& \multirow{2}{*}{\textbf{Avg}} \\
\cmidrule(lr){4-5} \cmidrule(lr){6-8} \cmidrule(lr){9-11}
 & & & CoLA & SST-2 & MRPC & STS-B & QQP & MNLI & QNLI & RTE & \\
\midrule
BERT \citep{devlin2019bert} & 110M & 512 & 59.0 & 93.1 & 89.5 & 89.4 & 91.4 & 85.4 & 91.6 & 78.2 & 84.84 \\
RoBERTa \citep{liu2019robertarobustlyoptimizedbert} & 125M & 512 & 63.6 & 94.8 & 90.2 & 91.2 & 91.9 & 87.6 & 92.8 & 78.7 & 86.35 \\
DeBERTav3 \citep{he2023debertav3improvingdebertausing} & 183M & 512 & \textbf{69.2} & 95.6 & 89.5 & 91.6 & \textbf{92.4} & \textbf{90.0} & \textbf{94.0} & 83.8 & 88.26 \\
MosaicBERT-128 \citep{portes2024mosaicbertbidirectionalencoderoptimized} & 137M & 128 & 58.2 & 93.5 & 89.0 & 90.3 & 92.0 & 85.6 & 91.4 & 83.0 & 85.38 \\
NomicBERT-2048 \citep{nussbaum2025nomicembedtrainingreproducible} & 137M & 2048 & 50.0 & 93.0 & 88.0 & 90.0 & 92.0 & 86.0 & 92.0 & 82.0 & 84.13 \\
GTE-en-MLM \citep{li2023generaltextembeddingsmultistage} & 137M & 8192 & 57.0 & 93.4 & 92.1 & 90.2 & 88.8 & 86.7 & 91.9 & 84.8 & 85.99 \\
ModernBERT \citep{warner2024modernbert} & 149M & 8192 & 65.1 & \textbf{96.0} & 92.2 & 91.8 & 92.1 & 89.1 & 93.9 &87.4 & 88.45 \\
\midrule
ModernALBERT-tiny & 50M & 1024 & 58.4 & 93.0 & 90.2 & 90.4 & 90.5 & 84.6 & 91.3  & 81.2 &  84.95 \\
ModernALBERT-medium & 55M & 1024 & 62.3 & 94.0 & 91.0 & 91.2 & 91.0 & 86.6 & 92.0 & 81.6 & 86.21\\
ModernALBERT-base & 75M & 1024 & 64.6 & 95.2 & 91.7 & 91.8 & 91.5 & 88.2 & 93.0 & 85.6 & 87.7 \\
ModernALBERT-large & 120M & 1024 & 66.4 & 95.5 & \textbf{92.7} & \textbf{92.1} & 92.0 & 88.9 & 93.7 & \textbf{88.44} & \textbf{88.72} \\
\bottomrule
\end{tabular}
}
\caption{GLUE benchmark results for base models. Scores are reported for each task category. \textbf{Avg} is the unweighted average across all tasks.}
\label{tab:glue}
\end{table*}

\section{Experimental Results}
\label{sec:results}

\begin{table}[h!]
\centering
\begin{tabular}{lcc}
\toprule
Model & F1 (\%) & Exact (\%) \\
\midrule
BERT-base & 88.6 & 80.6 \\
RoBERTa-base & 91.7 & 84.7 \\
ALBERT-xxlarge & 92.5 & 84.5 \\ 
ModernBERT-base & 92.6 & 85.2 \\
\midrule
\multicolumn{3}{c}{\textbf{Ours}} \\
\midrule
ModernALBERT-tiny & 90.0 & 82.9 \\
ModernALBERT-medium & 90.4 & 82.9 \\
ModernALBERT-base & 92.8 & \textbf{86.1} \\
ModernALBERT-large & \textbf{92.9} & 85.9 \\
\bottomrule
\end{tabular}
\caption{Results on the SQuAD-v2 Benchmark.}
\label{tab:squad2-results}
\end{table}

We evaluate ModernALBERT on natural language understanding, question answering, and information retrieval benchmarks. Our goal is to show that adding the Mixture-of-LoRAs (MoL) layer to a recursive, parameter-shared architecture provides substantial efficiency gains without sacrificing representational capacity. As shown in Table~\ref{tab:glue}, ModernALBERT performs strongly across model scales. 


\subsection{GLUE Benchmark Results}
We present GLUE benchmark results \citep{wang2019gluemultitaskbenchmarkanalysis} in Table~\ref{tab:glue}, where ModernALBERT performs strongly across model scales. The largest variant, ModernALBERT-large (120M parameters), achieves a GLUE average of \textbf{88.72}, surpassing the fully parameterised ModernBERT-base (149M parameters, 88.45) and outperforming previous compact models such as NomicBERT and MosaicBERT. The architectural benefits appear most clearly on tasks that require precise semantic matching and inference from limited data, where ModernALBERT-large achieves state-of-the-art results among base-class models on \textbf{RTE (88.44)}, \textbf{STS-B (92.1)}, and \textbf{MRPC (92.7)}.

For smaller datasets like CoLA and RTE, we further stabilise training by pre-training the router on MNLI and freezing its parameters. This allows the model to exploit a more robust, generalised routing policy, though we find the degree of improvement varies depending on the specific downstream objective. These optimisations preserve general capacity, keeping ModernALBERT competitive on high-resource tasks. On tasks such as MNLI, QNLI, and SST-2, it rivals substantially larger baselines, achieving parity with ModernBERT on QNLI (93.7) and MNLI (88.9). These results confirm that ModernALBERT successfully combines the efficiency of conditional computation with the robustness of fully parameterised transformers.

\paragraph{Question Answering Performance} We evaluate extractive question answering capabilities on SQuAD-v2 \citep{rajpurkar2018knowdontknowunanswerable}, as reported in Table~\ref{tab:squad2-results}. ModernALBERT-base achieves an F1 score of 92.8, outperforming the larger ModernBERT-base (92.6) and ALBERT-xxlarge (92.5). These results indicate that the Mixture-of-LoRAs approach transfers well to token-level classification tasks. ModernALBERT also reaches an Exact Match score of 86.1, showing that the model can localise answer spans accurately despite heavy parameter sharing.

\paragraph{Information Retrieval Results} Performance on the BEIR information retrieval benchmark \citep{thakur2021beirheterogenousbenchmarkzeroshot} is detailed in Table~\ref{tab:beir}. ModernALBERT shows strong domain generalisation, achieving the best average performance among compact models on tasks such as TREC-COVID, FiQA, and ArguAna. Specifically, on ArguAna (argument retrieval), ModernALBERT scores 48.82, outperforming ModernBERT (35.7). This indicates that our routing mechanism allows the model to adapt effectively to specialised domains and shifting distributions, a capability often lacking in static recursive transformers.

\begin{table*}[h!]
\centering
\resizebox{\textwidth}{!}{%
\begin{tabular}{lccccccc}
\toprule
Task & BERT & RoBERTa & DeBERTaV3 & NomicBERT & GTE-en & ModernBERT & ModernALBERT \\
\midrule
NFCorpus       & 24.3 & 20.4 & 8.0  & 25.7 & \textbf{26.3} & 23.7 & 24.30 \\
SciFact        & 51.3 & 45.6 & 22.6 & 52.0 & 54.1 & \textbf{57.0} & 56.90\\
TREC-Covid     & 49.5 & 52.2 & 48.4 & 63.0 & 49.7 & 72.1 & \textbf{72.85}\\
FiQA           & 22.8 & 26.1 & 11.5 & 23.5 & 30.1 & 28.8 & \textbf{30.43}\\
ArguAna        & 31.6 & 35.2 & 26.1 & 35.5 & 35.7 & 35.7 & \textbf{48.82}\\
Avg. (Subset)  & 38.9 & 37.7 & 20.2 & 41.0 & 41.4 & 41.6 & \textbf{46.66} \\
\bottomrule
\end{tabular}
}
\caption{Performance on selected BEIR benchmark datasets.}
\label{tab:beir}
\end{table*}

\subsection{Ablation Studies}
To isolate the contributions of our architectural choices, we conducted ablation studies comparing the MoL design against alternative conditional computation strategies and initialisation methods. A comparative analysis reveals that the proposed MoL design consistently outperforms the Mixture-of-Adapters (MoA) approach. In the MoA configuration, a set of $n$ adapters and a router are placed sequentially \textit{after} the feed-forward network, whereas our MoL design integrates LoRA experts directly into the weight space of the shared FFN. As shown in Table~\ref{tab:ablation_moa}, with 8 experts and top-2 routing, MoL achieves a GLUE average of 77.24, compared to 76.87 for the equivalent MoA configuration. This performance gap suggests that modifying the internal representations of the shared FFN via LoRA facilitates a deeper and more effective integration of conditional computation than merely appending adaptive layers to the block output.

Table~\ref{tab:ablation_moa} further illustrates the impact of scaling the expert count. Increasing the capacity from a single LoRA expert (Average: 76.08) to 8 experts with top-2 routing results in a clear performance gain of 1.16 points. This trend confirms that the observed improvements are driven by the conditional capacity inherent in the mixture model rather than the simple addition of trainable parameters. Finally, we compare our ModernALBERT model with the Relaxed Recursive Transformer (RRT) baseline using the same step-wise initialisation strategy proposed in RRT and training both models with the masked language modelling (MLM) objective. As shown in Table~\ref{tab:init-ablation}, ModernALBERT achieves a higher GLUE average (\textbf{81.94 vs. 80.95}), indicating that integrating conditional computation directly into the shared FFN weights yields more effective training than the relaxed recursive approach.

The training configuration was kept uniform across all ablation experiments (including MoA, MoL scaling, and RRT comparisons), all models were trained for a consistent $30$k steps on the Wikipedia corpus, using a batch size of $128$ and a maximum sequence length of $1024$.

\begin{table*}[h]
\centering
\resizebox{\textwidth}{!}{%
\begin{tabular}{lccc
                cc
                cc
                cc
                c}
\toprule
\multirow{2}{*}{\textbf{Model}} & \multirow{2}{*}{\textbf{Params}} & \multirow{2}{*}{\textbf{Num Experts}} & \multirow{2}{*}{\textbf{Top-K}} 
& \multicolumn{2}{c}{\textbf{Single Sentence}} 
& \multicolumn{2}{c}{\textbf{Paraphrase}} 
& \multicolumn{2}{c}{\textbf{NLI}} 
& \multirow{2}{*}{\textbf{Avg}} \\
\cmidrule(lr){5-6} \cmidrule(lr){7-8} \cmidrule(lr){9-10}
 & & & & CoLA & SST-2 & MRPC & STS-B & MNLI & RTE & \\
\midrule
ModernBERT & 245M &  -  & - & 34.05 & 88.76 & 88.72 & 88.22 & 81.53 & 75.45 & 76.12 \\
ModernALBERT (Base) & 56M & - & - & 37.39 & 88.76 & 86.76 & 87.40 & 81.14 & 72.92 & 75.73 \\
\midrule
\multicolumn{11}{c}{\textit{Mixture of Adapters (MoA) - Adapters placed after FFN}} \\
MoA & 64M & 1 & 1 & 38.49 & 88.64 & 87.25 & 87.68 & 80.77 & 71.84 & 75.45 \\
MoA & 69M & 4 & 1 & 38.49 & \textbf{90.71} & 87.50 & 87.32 & 80.61 & 74.36 & 76.17 \\
MoA & 75M & 8 & 2 & \underline{41.80} & 89.10 & 87.74 & 87.56 & 81.05 & 74.00 & \underline{76.87}\\
\midrule
\multicolumn{11}{c}{\textit{Mixture of LoRAs (MoL) - Proposed Method}} \\
MoL & 73M & 1 & 1 & 40.06 & 88.99 & 87.74 & 87.38 & 81.03 & 73.28 & 76.08 \\
MoL & 74M & 4 & 1 & 40.40 & 89.00 & 87.99 & 87.33 & 80.91 & 73.40 &  76.50 \\
MoL & 75M & 8 & 2 & \textbf{42.97} & \underline{90.36} & \underline{88.23} & \underline{88.01} & \underline{81.15} & \underline{74.72} & \textbf{77.24} \\
\bottomrule
\end{tabular}
}
\caption{Ablation Study: Mixture of Adapters (MoA) vs. Mixture of LoRAs (MoL). MoL consistently outperforms MoA at similar parameter counts.}
\label{tab:ablation_moa}
\end{table*}

\begin{table*}[h!]
\centering
\resizebox{\textwidth}{!}{%
\begin{tabular}{lc cc ccc ccc c}
\toprule
\multirow{2}{*}{\textbf{Model}} & \multirow{2}{*}{\textbf{Params}} 
& \multicolumn{2}{c}{\textbf{Single Sentence}} 
& \multicolumn{3}{c}{\textbf{Paraphrase/Similarity}} 
& \multicolumn{3}{c}{\textbf{Natural Language Inference}} 
& \multirow{2}{*}{\textbf{Avg}} \\
\cmidrule(lr){3-4}\cmidrule(lr){5-7}\cmidrule(lr){8-10}
 & & CoLA & SST-2 & MRPC & STS-B & QQP & MNLI & RTE & QNLI & \\
\midrule
ModernALBERT & 75M 
& \textbf{45.69} & 89.90 
& \textbf{88.72} & \textbf{89.60} & \textbf{90.86} 
& \textbf{83.74} & \textbf{76.17} & \textbf{90.82} 
& \textbf{81.94} \\
Relaxed Recursive Transformers & 73M 
& 40.76 & \textbf{90.02} 
& 87.99 & 88.99 & 90.86 
& 83.25 & 75.09 & 90.66 
& 80.95 \\
\bottomrule
\end{tabular}
}
\caption{Ablation on Initialisation: We use the same initialisation as Relaxed Recursive Transformers (RRT) and compare our Mixture-of-LoRAs (MoL) strategy with their layer-wise LoRAs on GLUE tasks.}
\label{tab:init-ablation}
\end{table*}

\section{Optimising MoL for Inference}

The proposed MoL layer augments a parameter-shared recursive transformer with lightweight LoRA experts coordinated by a routing network. While this design captures the flexibility of Mixture-of-Experts architectures, the conditional routing mechanism introduces latency overhead during inference by evaluating multiple experts per token. To address this, we propose compressing the MoL layer into a single static adapter for deployment.

\begin{table*}[t!]
\centering
\begin{tabular}{lccc}
\toprule
Model & Latency (ms) $\downarrow$ & Throughput (token/s) $\uparrow$ & Memory (GB) $\downarrow$ \\
\midrule
ModernBERT-base & 14.44 & 69519 & 0.570\\
ModernBERT-large & 21.30 & 48117 & 1.5\\
\midrule
ModernALBERT-tiny & \textbf{9.46} & \textbf{106527} & \textbf{0.196} \\
ModernALBERT-medium & 9.72 & 106208 & 0.207 \\
ModernALBERT-base & 12.71 & 80731 & 0.291 \\
ModernALBERT-large & 18.87 & 54483 & 0.459 \\
\bottomrule
\end{tabular}
\caption{Efficiency Metrics of Different Models when Using Merged Experts.}
\label{tab:efficiency}
\end{table*}

\subsection{Expert Merging Strategies}

To eliminate conditional branching, we synthesise a single dense adapter $\Delta_{\text{merged}}$ from the pool of $E$ experts $\{\Delta_1, \dots, \Delta_E\}$. This aggregation is defined by a weighting vector $\mathbf{w} \in \mathbb{R}^E$:
\begin{align}
    \Delta_{\text{merged}} = \sum_{j=1}^E w_j \Delta_j.
\end{align}
During fine-tuning, the dynamic routing mechanism is disabled, and the model updates $\Delta_{\text{merged}}$ directly \citep{he2023mergingexpertsoneimproving}. This procedure effectively converts the sparse MoL layer into a standard, static LoRA module, reducing the inference cost to that of a dense model. As shown in Table~\ref{tab:efficiency}, this approach yields substantial efficiency gains; ModernALBERT-tiny achieves 9.46 ms latency and 106{,}527 tokens/s throughput while using only 0.196 GB of GPU memory.

We investigate two strategies for managing this aggregation during fine-tuning.

\subsubsection{Uniform Initialization}
In the first strategy, we initialise the merged adapter by averaging the pre-trained experts uniformly ($w_j = 1/E$). At each step, the merged adapter $\Delta_{\text{merged}}$ is formed by averaging the experts, and the model parameters are updated via standard backpropagation. This method assumes a neutral starting point, allowing the aggregate representation to adapt to the downstream task without reliance on token-specific routing.

\subsubsection{Dynamic EMA Merging}
The second strategy updates the merging weights $\mathbf{w}$ dynamically throughout fine-tuning based on the router's activation patterns. We initialise $\mathbf{w}$ uniformly ($w_j = 1/E$), allowing the EMA updates to gradually refine the distribution from a neutral starting point. Let $p_{i,t} \in \mathbb{R}^E$ be the router probability vector for token $t$ in sample $i$. We first compute a per-sample routing vector $r_i$ and a batch-level average $r_b$:
\begin{align}
    r_i = \frac{1}{T_i}\sum_{t=1}^{T_i} p_{i,t}, \quad \quad r_b = \frac{1}{B}\sum_{i=1}^{B} r_i.
\end{align}
We then update the global merging weights $\mathbf{w}$ using an exponential moving average (EMA) of these batch statistics:
\begin{align}
    \mathbf{w}_{\text{new}} = \alpha \mathbf{w}_{\text{old}} + (1-\alpha) r_b,
\end{align}
where $\alpha$ is a decay factor. Regarding the router itself, we adopt a task-dependent approach: for data-rich tasks like MNLI, the router is trained jointly to refine expert selection; for smaller datasets (STS-B, RTE, SST-2), the router is frozen to leverage pre-trained specialisation while only the EMA weights adapt.

As shown in Table~\ref{tab:merged-experts}, the EMA-based strategy consistently outperforms uniform initialization (Vanilla). By dynamically adjusting the expert contributions based on routing statistics, EMA preserves nearly all of the unmerged model's accuracy, with particularly strong gains on RTE (86.28 vs 84.83) and SST-2. These results confirm that incorporating the router's learned preferences into the merging process, even when the router itself is frozen, provides a superior adaptation mechanism compared to a static uniform initialization.

\section{Discussion}

The increase in capacity in the MoL architecture arises primarily from expert specialisation. As the router learns to activate different LoRA modules for different inputs, each module becomes responsible for modelling particular aspects of the data distribution, such as syntactic cues, semantic relations, or domain-specific patterns. This division of responsibility yields richer internal representations and helps mitigate the expressivity limitations inherent to fully shared architectures. Although the additional experts provide clear performance gains, dynamic routing naturally increases inference cost because several experts may be activated per token \citep{shazeer2017mixture, lepikhin2020gshard, fedus2021switch}.

Our results on expert merging further suggest that while dynamic routing is crucial for training high-capacity models, the learned expressivity can often be compressed into a static representation for inference. The success of the EMA strategy implies that the aggregate behaviour of the experts captures a robust policy that adapts well across general domains. This finding aligns with recent work on model merging and distillation, indicating that the computational cost of conditional branching is beneficial for learning but not always strictly necessary for maintaining performance during deployment. This trade-off allows MoL-based models to occupy a unique position where they train with the expressivity of large MoEs but deploy with the speed and footprint of compact dense models.

\section{Limitations}
While the proposed model improves parameter efficiency and adaptive capacity through the integration of MoL layers, several limitations remain. First, despite the architectural optimisations and merging strategies explored in this work, the computational complexity of mixture-of-experts routing remains relatively high compared to fully shared or purely dense alternatives. Although our design mitigates dispatching overhead through lightweight routing and optional merging, further advances in efficient expert selection and load balancing are necessary to fully realise the scalability benefits of conditional computation.

Second, the model employs global attention across all tokens without incorporating locality-aware mechanisms. Recent architectures such as ModernBERT have demonstrated that local or block-sparse attention can substantially improve both efficiency and long-sequence modelling performance. Consequently, our approach may underperform on tasks requiring extended context retention or fine-grained reasoning over long inputs. Future work could address these issues by integrating local attention patterns or hybrid attention schemes within the recursive parameter-sharing framework.

\section{Conclusion and Future Work}
\begin{table}[t!]
\centering
\begin{tabular}{lcccc}
\toprule
Method & MNLI & STS-B & RTE & SST-2 \\
\midrule
No Merging & 88.2 & 91.8 & 85.6 & 95.2 \\
Vanilla & 87.9 & 91.5 & 84.83 & 94.8 \\
EMA & \textbf{88.0} & \textbf{91.7} & \textbf{86.28} & \textbf{95.1} \\
\bottomrule
\end{tabular}
\caption{Effect of Expert Merging on Different Datasets.}
\label{tab:merged-experts}
\end{table}

In this work, we explored several strategies to enhance the representational power of recursive transformers. Building upon the principle of parameter sharing, we introduced the Mixture-of-LoRAs (MoL) layer, a novel mechanism that integrates conditional computation into the shared transformer architecture. The proposed layer demonstrated substantial improvements in performance while maintaining parameter efficiency, highlighting its effectiveness as a scalable and adaptable extension to recursive parameter-shared models.

For future research, extending this framework to multimodal settings offers a promising direction for studying conditional representations across diverse input types. Moreover, applying parameter sharing in conjunction with the proposed Mixture-of-LoRAs to large autoregressive language models could provide a path toward reducing their parameter count and memory footprint, while preserving or even enhancing their representational capacity. Such integration may enable the development of more efficient yet powerful large-scale generative models.

\section*{Acknowledgments}
This research was supported in part by Lambda, Inc.\ through its Research Grant Program, and by a micro-grant from Trelis Research.

\nocite{langley00}

\bibliography{example_paper}

@misc{sanh2020distilbertdistilledversionbert,
      title={DistilBERT, a distilled version of BERT: smaller, faster, cheaper and lighter}, 
      author={Victor Sanh and Lysandre Debut and Julien Chaumond and Thomas Wolf},
      year={2020},
      eprint={1910.01108},
      archivePrefix={arXiv},
      primaryClass={cs.CL},
      url={https://arxiv.org/abs/1910.01108}, 
}

@misc{rajpurkar2018knowdontknowunanswerable,
      title={Know What You Don't Know: Unanswerable Questions for SQuAD}, 
      author={Pranav Rajpurkar and Robin Jia and Percy Liang},
      year={2018},
      eprint={1806.03822},
      archivePrefix={arXiv},
      primaryClass={cs.CL},
      url={https://arxiv.org/abs/1806.03822}, 
}

@misc{hinton2015distillingknowledgeneuralnetwork,
      title={Distilling the Knowledge in a Neural Network}, 
      author={Geoffrey Hinton and Oriol Vinyals and Jeff Dean},
      year={2015},
      eprint={1503.02531},
      archivePrefix={arXiv},
      primaryClass={stat.ML},
      url={https://arxiv.org/abs/1503.02531}, 
}

@misc{kingma2017adammethodstochasticoptimization,
      title={Adam: A Method for Stochastic Optimization}, 
      author={Diederik P. Kingma and Jimmy Ba},
      year={2017},
      eprint={1412.6980},
      archivePrefix={arXiv},
      primaryClass={cs.LG},
      url={https://arxiv.org/abs/1412.6980}, 
}

@inproceedings{together2023redpajamas,
  title={RedPajama: An Open Dataset for Training Large Language Models},
  author={Computer, Together},
  booktitle={Thirty-eighth Conference on Neural Information Processing Systems Datasets and Benchmarks Track},
  year={2023},
  url={https://github.com/togethercomputer/RedPajama-Data}
}

@misc{penedo2023refinedwebdatasetfalconllm,
      title={The RefinedWeb Dataset for Falcon LLM: Outperforming Curated Corpora with Web Data, and Web Data Only}, 
      author={Guilherme Penedo and Quentin Malartic and Daniel Hesslow and Ruxandra Cojocaru and Alessandro Cappelli and Hamza Alobeidli and Baptiste Pannier and Ebtesam Almazrouei and Julien Launay},
      year={2023},
      eprint={2306.01116},
      archivePrefix={arXiv},
      primaryClass={cs.CL},
      url={https://arxiv.org/abs/2306.01116}, 
}

@article{nguyen2019transformers,
  title={Transformers without tears: Improving the normalization of self-attention},
  author={Nguyen, Toan Q and Salazar, Julian},
  journal={arXiv preprint arXiv:1910.05895},
  year={2019}
}

@misc{thakur2021beirheterogenousbenchmarkzeroshot,
      title={BEIR: A Heterogenous Benchmark for Zero-shot Evaluation of Information Retrieval Models}, 
      author={Nandan Thakur and Nils Reimers and Andreas Rücklé and Abhishek Srivastava and Iryna Gurevych},
      year={2021},
      eprint={2104.08663},
      archivePrefix={arXiv},
      primaryClass={cs.IR},
      url={https://arxiv.org/abs/2104.08663}, 
}

@article{warner2024modernbert,
  author    = {Benjamin Warner and Antoine Chaffin and Benjamin Clavi{\'e} and Orion Weller and others},
  title     = {ModernBERT: A Modern Bidirectional Encoder for Fast, Memory-Efficient, and Long-Context Fine-tuning and Inference},
  journal   = {arXiv preprint},
  year      = {2024},
  url       = {https://arxiv.org/abs/2412.13663}
}

@misc{wang2019gluemultitaskbenchmarkanalysis,
      title={GLUE: A Multi-Task Benchmark and Analysis Platform for Natural Language Understanding}, 
      author={Alex Wang and Amanpreet Singh and Julian Michael and Felix Hill and Omer Levy and Samuel R. Bowman},
      year={2019},
      eprint={1804.07461},
      archivePrefix={arXiv},
      primaryClass={cs.CL},
      url={https://arxiv.org/abs/1804.07461}, 
}

@misc{houlsby2019parameterefficienttransferlearningnlp,
      title={Parameter-Efficient Transfer Learning for NLP}, 
      author={Neil Houlsby and Andrei Giurgiu and Stanislaw Jastrzebski and Bruna Morrone and Quentin de Laroussilhe and Andrea Gesmundo and Mona Attariyan and Sylvain Gelly},
      year={2019},
      eprint={1902.00751},
      archivePrefix={arXiv},
      primaryClass={cs.LG},
      url={https://arxiv.org/abs/1902.00751}, 
}

@inproceedings{vaswani2017attention,
  author    = {Ashish Vaswani and Noam Shazeer and Niki Parmar and Jakob Uszkoreit and
               Llion Jones and Aidan N. Gomez and {\L}ukasz Kaiser and Illia Polosukhin},
  title     = {Attention is All You Need},
  booktitle = {Advances in Neural Information Processing Systems (NeurIPS)},
  year      = {2017},
  url       = {https://arxiv.org/abs/1706.03762}
}

@inproceedings{devlin2019bert,
  author    = {Jacob Devlin and Ming{-}Wei Chang and Kenton Lee and Kristina Toutanova},
  title     = {BERT: Pre-training of Deep Bidirectional Transformers for Language Understanding},
  booktitle = {Proceedings of NAACL-HLT},
  year      = {2019},
  url       = {https://arxiv.org/abs/1810.04805}
}

@inproceedings{nouriborji2022minialbert,
  author    = {Mohammadmahdi Nouriborji and others},
  title     = {MiniALBERT: Model Distillation via Parameter-Efficient Recursive Student},
  booktitle = {EACL / ACL (workshop / proceedings)},
  year      = {2023},
  url       = {https://aclanthology.org/2023.eacl-main.83.pdf}
}

@article{brown2020language,
  author    = {Tom B. Brown and Benjamin Mann and Nick Ryder and Melanie Subbiah and Jared Kaplan
               and Prafulla Dhariwal and Arvind Neelakantan and Pranav Shyam and Girish Sastry
               and Amanda Askell and Sandhini Agarwal and Ariel Herbert-Voss and Gretchen Krueger
               and Tom Henighan and Rewon Child and Aditya Ramesh and Daniel M. Ziegler and Jeffrey Wu
               and Clemens Winter and Christopher Hesse and Mark Chen and Eric Sigler and Mateusz Litwin
               and Scott Gray and Benjamin Chess and Jack Clark and Christopher Berner and Sam McCandlish
               and Alec Radford and Ilya Sutskever and Dario Amodei},
  title     = {Language Models are Few-Shot Learners},
  journal   = {Advances in Neural Information Processing Systems (NeurIPS) / arXiv preprint},
  year      = {2020},
  url       = {https://arxiv.org/abs/2005.14165}
}

@misc{he2023mergingexpertsoneimproving,
      title={Merging Experts into One: Improving Computational Efficiency of Mixture of Experts}, 
      author={Shwai He and Run-Ze Fan and Liang Ding and Li Shen and Tianyi Zhou and Dacheng Tao},
      year={2023},
      eprint={2310.09832},
      archivePrefix={arXiv},
      primaryClass={cs.CL},
      url={https://arxiv.org/abs/2310.09832}, 
}

@article{lan2019albert,
  author    = {Zhenzhong Lan and Mingda Chen and Sebastian Goodman and Kevin Gimpel and Piyush Sharma and Radu Soricut},
  title     = {ALBERT: A Lite BERT for Self-supervised Learning of Language Representations},
  year      = {2019},
  note      = {arXiv preprint},
  url       = {https://arxiv.org/abs/1909.11942}
}

@article{dehghani2018universal,
  author    = {Mostafa Dehghani and Stephan Gouws and Oriol Vinyals and Jakob Uszkoreit and Lukasz Kaiser},
  title     = {Universal Transformers},
  year      = {2018},
  note      = {arXiv preprint},
  url       = {https://arxiv.org/abs/1807.03819}
}

@article{shazeer2017mixture,
  author    = {Noam Shazeer and Azalia Mirhoseini and Krzysztof Maziarz and Andy Davis and Quoc V. Le and Geoffrey E. Hinton and Jeff Dean},
  title     = {Outrageously Large Neural Networks: The Sparsely-Gated Mixture-of-Experts Layer},
  year      = {2017},
  note      = {ICLR workshop / arXiv preprint},
  url       = {https://arxiv.org/abs/1701.06538}
}

@article{fedus2021switch,
  author    = {William Fedus and Barret Zoph and Noam Shazeer},
  title     = {Switch Transformers: Scaling to Trillion-Parameter Models with Simple and Efficient Sparsity},
  year      = {2021},
  note      = {arXiv / conference},
  url       = {https://arxiv.org/abs/2101.03961}
}

@article{lepikhin2020gshard,
  author    = {Dmitry Lepikhin and HyoukJoong Lee and Yuanzhong Xu and Dehao Chen and Orhan Firat and Yanping Huang and Maxim Krikun and Noam Shazeer and Zhifeng Chen},
  title     = {GShard: Scaling Giant Models with Conditional Computation and Automatic Sharding},
  year      = {2020},
  note      = {arXiv preprint / ICLR},
  url       = {https://arxiv.org/abs/2006.16668}
}

@article{hu2021lora,
  author    = {Edward J. Hu and Yelong Shen and Phillip Wallis and Zeyuan Allen{-}Zhu and Yuanzhi Li and Shean Wang and Weizhu Chen},
  title     = {LoRA: Low-Rank Adaptation of Large Language Models},
  year      = {2021},
  note      = {arXiv preprint},
  url       = {https://arxiv.org/abs/2106.09685}
}

@misc{portes2024mosaicbertbidirectionalencoderoptimized,
      title={MosaicBERT: A Bidirectional Encoder Optimized for Fast Pretraining}, 
      author={Jacob Portes and Alex Trott and Sam Havens and Daniel King and Abhinav Venigalla and Moin Nadeem and Nikhil Sardana and Daya Khudia and Jonathan Frankle},
      year={2024},
      eprint={2312.17482},
      archivePrefix={arXiv},
      primaryClass={cs.CL},
      url={https://arxiv.org/abs/2312.17482}, 
}

@misc{he2023debertav3improvingdebertausing,
      title={DeBERTaV3: Improving DeBERTa using ELECTRA-Style Pre-Training with Gradient-Disentangled Embedding Sharing}, 
      author={Pengcheng He and Jianfeng Gao and Weizhu Chen},
      year={2023},
      eprint={2111.09543},
      archivePrefix={arXiv},
      primaryClass={cs.CL},
      url={https://arxiv.org/abs/2111.09543}, 
}

@misc{liu2019robertarobustlyoptimizedbert,
      title={RoBERTa: A Robustly Optimized BERT Pretraining Approach}, 
      author={Yinhan Liu and Myle Ott and Naman Goyal and Jingfei Du and Mandar Joshi and Danqi Chen and Omer Levy and Mike Lewis and Luke Zettlemoyer and Veselin Stoyanov},
      year={2019},
      eprint={1907.11692},
      archivePrefix={arXiv},
      primaryClass={cs.CL},
      url={https://arxiv.org/abs/1907.11692}, 
}

@misc{li2023generaltextembeddingsmultistage,
      title={Towards General Text Embeddings with Multi-stage Contrastive Learning}, 
      author={Zehan Li and Xin Zhang and Yanzhao Zhang and Dingkun Long and Pengjun Xie and Meishan Zhang},
      year={2023},
      eprint={2308.03281},
      archivePrefix={arXiv},
      primaryClass={cs.CL},
      url={https://arxiv.org/abs/2308.03281}, 
}

@misc{nussbaum2025nomicembedtrainingreproducible,
      title={Nomic Embed: Training a Reproducible Long Context Text Embedder}, 
      author={Zach Nussbaum and John X. Morris and Brandon Duderstadt and Andriy Mulyar},
      year={2025},
      eprint={2402.01613},
      archivePrefix={arXiv},
      primaryClass={cs.CL},
      url={https://arxiv.org/abs/2402.01613}, 
}

@article{shazeer2020glu,
  author    = {Noam Shazeer},
  title     = {GLU Variants Improve Transformer},
  year      = {2020},
  note      = {arXiv preprint (describes GeGLU, SwiGLU, etc.)},
  url       = {https://arxiv.org/abs/2002.05202}
}

@article{su2021roformer,
  author    = {Jianlin Su and Yu Lu and Shengfeng Pan and Ahmed Murtadha and Bo Wen and Yunfeng Liu},
  title     = {RoFormer: Enhanced Transformer with Rotary Position Embedding},
  year      = {2021},
  note      = {arXiv preprint},
  url       = {https://arxiv.org/abs/2104.09864}
}

@inproceedings{dao2022flashattention,
  author    = {Tri Dao and Daniel Y. Fu and Stefano Ermon and Atri Rudra and Christopher R\'e},
  title     = {FlashAttention: Fast and Memory-Efficient Exact Attention with IO-Awareness},
  booktitle = {Advances in Neural Information Processing Systems (NeurIPS)},
  year      = {2022},
  url       = {https://arxiv.org/abs/2205.14135}
}

@article{bae2024effective,
  author    = {Seongjin Bae and others},
  title     = {Effective Parameter Sharing with Layer-wise LoRA},
  year      = {2024},
  note      = {introduces Relaxed Recursive Transformers / layer-wise LoRA (openreview/arXiv)},
  url       = {https://arxiv.org/abs/2410.20672}
}

@article{mixlora2024,
  title   = {MixLoRA: Enhancing Large Language Models Fine-Tuning with LoRA-based Mixture of Experts},
  author  = {D. Li and Y. Ma and N. Wang and others},
  year    = {2024},
  journal = {arXiv preprint},
  url     = {https://arxiv.org/abs/2404.15159}
}

@article{mole2024,
  title   = {Mixture of LoRA Experts},
  author  = {Xun Wu and Shaohan Huang and Furu Wei and others},
  year    = {2024},
  journal = {arXiv preprint / OpenReview},
  url     = {https://arxiv.org/abs/2404.13628}
}

@article{moa2024,
  title   = {Mixture-of-LoRAs: An Efficient Multitask Tuning for Large Language Models},
  author  = {W. Feng and others},
  year    = {2024},
  journal = {arXiv preprint},
  url     = {https://arxiv.org/abs/2403.03432}
}

@article{adapterfusion2020,
  title   = {Non-Destructive Task Composition for Transfer Learning (AdapterFusion)},
  author  = {Jonas Pfeiffer and Andreas Rücklé and Klaus Sch{\"u}tze and others},
  year    = {2020},
  journal = {arXiv preprint / ACL/EACL workshop},
  url     = {https://arxiv.org/abs/2005.00247}
}

@article{adaptermix2023,
  title   = {ADAPTERMIX: Exploring the Efficacy of Mixture-of-Adapters},
  author  = {A. Mehrish and others},
  year    = {2023},
  journal = {arXiv preprint},
  url     = {https://arxiv.org/abs/2305.18028}
}
\bibliographystyle{icml2025}




\end{document}